\pdfoutput=1
\documentclass[11pt]{article}

\usepackage[final]{acl}

\usepackage{times}
\usepackage{latexsym}

\usepackage[T1]{fontenc}
\usepackage[utf8]{inputenc}

\usepackage{microtype}

\usepackage{inconsolata}

\usepackage{graphicx}

\usepackage{hyperref}       % hyperlinks
\usepackage{url}            % simple URL typesetting
\usepackage{booktabs}       % professional-quality tables
\usepackage{amsfonts}       % blackboard math symbols
\usepackage{nicefrac}       % compact symbols for 1/2, etc.
\usepackage{microtype}      % microtypography
\usepackage{multirow}
\usepackage{amssymb}
\usepackage{amsbsy}
\usepackage{amsmath}
\usepackage{amsthm}
\usepackage{latexsym}
\usepackage{subcaption}
\usepackage{wrapfig}
\usepackage{float}
\usepackage{graphicx}
\usepackage{tabularx}
\usepackage{ulem}
\usepackage[font=small]{caption}
\usepackage{algorithm}
\usepackage{algorithmicx, algpseudocode}
\usepackage{mathtools}
\usepackage{arydshln}
\usepackage{titlesec}
\usepackage{booktabs}
\usepackage{xcolor}         % colors
\definecolor{myblue}{RGB}{33, 48, 163}
\definecolor{myred}{RGB}{192, 0, 0}
\usepackage{CJKutf8}
\usepackage{graphicx}
\usepackage{makecell}
\usepackage{nicefrac}
\usepackage{CJKutf8}
\usepackage{relsize} % 

\makeatletter
\renewcommand{\maketag@@@}[1]{\hbox{\m@th\normalsize\normalfont#1}}%
\makeatother

\title{Unilaw-R1: A Large Language Model for Legal Reasoning with Reinforcement Learning and Iterative Inference}

\author{
 \textbf{Hua Cai\textsuperscript{1 $\dagger$}},
 \textbf{Shuang Zhao\textsuperscript{1}},
 \textbf{Liang Zhang\textsuperscript{1}},
 \textbf{Xuli Shen\textsuperscript{1,2 $\dagger$}},
 \textbf{Qing Xu\textsuperscript{1}},
  \\
 \textbf{Weilin Shen\textsuperscript{1}},
 \textbf{Zihao Wen\textsuperscript{2}},
 \textbf{Tianke Ban\textsuperscript{2 $\dagger$ }}
\\
 \textsuperscript{1}UniDT,
 \textsuperscript{2}Fudan University
}

\begin{document}
\begin{CJK}{UTF8}{gbsn} 
\maketitle
\begin{abstract}

Reasoning-focused large language models (LLMs) are rapidly evolving across various domains, yet their capabilities in handling complex legal problems remains underexplored. In this paper, we introduce Unilaw-R1, a large language model tailored for legal reasoning. With a lightweight 7-billion parameter scale, Unilaw-R1 significantly reduces deployment cost while effectively tackling three core challenges in the legal domain: insufficient legal knowledge, unreliable reasoning logic, and weak business generalization. To address these issues, we first construct Unilaw-R1-Data, a high-quality dataset containing $\sim$17K distilled and screened chain-of-thought (CoT) samples. Based on this, we adopt a two-stage training strategy combining Supervised Fine-Tuning (SFT) and Reinforcement Learning (RL), which significantly boosts the model’s performance on complex legal reasoning tasks and supports interpretable decision-making in legal AI applications. To assess legal reasoning ability, we also introduce Unilaw-R1-Eval, a dedicated benchmark designed to evaluate models across single- and multi-choice legal tasks. Unilaw-R1 demonstrates strong results on authoritative benchmarks, outperforming all models of similar scale and achieving performance on par with the much larger DeepSeek-R1-Distill-Qwen-32B (54.9\%). Following domain-specific training, it also showed significant gains on LawBench and LexEval, exceeding Qwen-2.5-7B-Instruct (46.6\%) by an average margin of 6.6\%. Code is available at: https://github.com/Hanscal/Unilaw-R1.
{\let\thefootnote\relax\footnote{{{   $^{\dagger}$Corresponding authors: Tianke Ban (bantianke@fudan.\\edu.cn), Hua Cai (hua.cai@unidt.com),  Xuli Shen (xlshen20@fudan.edu.cn). }}}}
\end{abstract}

\section{Introduction}

In recent years, the rapid iteration of large language models (LLMs) has significantly propelled the evolution of artificial intelligence towards artificial general intelligence (AGI). Models such as OpenAI's o1-series \citep{openai2024llms} have enhanced their ability for complex reasoning tasks by extending the length of the "chain-of-thought" through an "exploration-reflection-iteration" mechanism. Similar o1-like LLMs, such as QwQ \citep{qwen} and Marco-o1 \citep{zhao2024marco}, have demonstrated significant improvements across tasks like mathematics, programming, and logical reasoning. 
% \footnote{EMNLP 2025 Accepted.}
Although general reasoning models exhibit considerable potential, their application in specialized domains such as legal is limited. Legal reasoning requires not only legal, economic and mathematical knowledge, but also step-by-step and verifiable logic. Existing models face three major challenges: (1) inconsistencies in legal data increase preprocessing complexity and weaken reasoning  \citep{koenecke2025tasks,mishra2025investigating,sheik2024neural,steging2023improving,aumiller2021structural};
(2) the black-box nature of LLMs lack transparency, falling short of traceability standards \citep{wang2023alpha,zhao2024explainability,tong2024ploutos,chaudhary2024unveiling};
and (3) insufficient legal knowledge leads to unreliable or incoherent reasoning processes \citep{blair2025llms,dahl2024large}.
Moreover, effective legal reasoning must adhere to both the external validity of codified law and the internal procedural consistency that ensures fairness and predictability in legal interpretation \citep{zou2021nine,raz2009authority,fuller1969morality}.
% (3) current models may generate inconsistent or uncertain reasoning paths when lacking adequate legal knowledge, which can compromise reliability in critical legal decision-making \citep{blair2025llms,dahl2024large}.

To address these, we introduce Unilaw-R1, a legal reasoning LLM built upon a high-quality legal dataset and optimized through a two-stage training paradigm. Unilaw-R1 overcomes fragmentation, opacity, and generalization issues in legal AI systems. Our key contributions are as follows:

\begin{itemize}
\item \textbf{High-Quality Legal Reasoning and Eval Dataset}: We propose Unilaw-R1-Data and Unilaw-R1-Eval datasets that constructed from multiple-choice questions. These cover a wide range of legal topics including civil law, criminal law, administrative law, and procedural law, providing a robust foundation for training and evaluation in legal scenarios. 
\item \textbf{Two-Stage Model Construction Framework}: We introduce a two-stage pipeline involving Supervised Fine-Tuning (SFT) on high-quality Chain-of-Thought (CoT) reasoning data, followed by Reinforcement Learning (RL) with a legal validity reward function integrated into GRPO. This design improves both reasoning accuracy and legal conformity.
\item \textbf{Explicit Legal Iterative Inference}: Unilaw-R1 incorporates an iterative multi-agent inference strategy, enabling advanced legal decision-making and strong generalization across diverse legal domains.

\end{itemize}

\begin{figure*}[ht] % figure 
    \centering % 
    \includegraphics[width=0.85\textwidth]{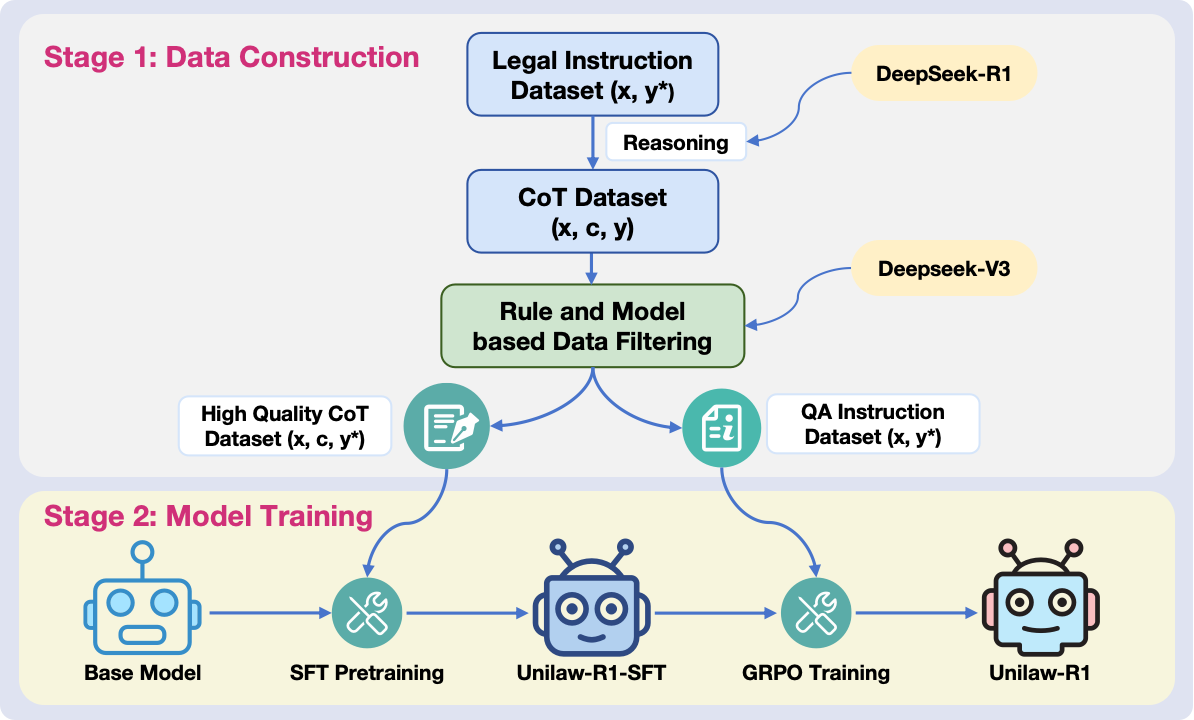}
    \caption{The pipeline for constructing Unilaw-R1. The diagram depicts the two-stage construction framework of Unilaw-R1: Data Generation (using DeepSeek-R1 for reasoning to generate CoT data, followed by quality filtering with the DeepSeek-V3) and Model Training (including SFT pretraining and GRPO optimization for Unilaw-R1).} % 
    \label{2stage} % 
\end{figure*}

\section{Related Work}
The capabilities of large language models have advanced rapidly through innovations in training paradigms and reasoning strategies. The o1-series models \citep{jaech2024openai} introduced iterative "exploration-reflection" mechanisms that lengthen the CoT process, thereby improving reasoning depth. Subsequent efforts such as QwQ \citep{qwen}, Marco-o1 \citep{zhao2024marco} and Fin-R1 \citep{liu2025fin} extended this approach across domains including logic, mathematics, and finance.
In the legal domain, compliant adaptations of o1-class models, such as HK-O1aw and PatientSeek \citep{HK-O1aw,PatientSeek}, have shown the potential of LLMs in simulating human-like legal reasoning. Yu et al. \cite{yu2025evaluating} further pushed this frontier by employing test-time scaling techniques to enhance performance on legal tasks.

Distinct from the above, DeepSeek-R1 \citep{guo2025deepseek} takes an efficient reinforcement learning (RL) approach, training LLMs via thousands of steps of unsupervised RL combined with a cold-start corpus and multi-stage curriculum learning. This strategy results in emergent reasoning capabilities and improved readability, highlighting the promise of RL in scaling inference power.

Despite significant advancements, applying LLMs to the legal domain introduces unique challenges due to domain-specific constraints. Previous research has emphasized the necessity for structured legal datasets, transparency, and reliable performance across scenarios, areas where current models still fall short. Unilaw-R1 addresses these gaps through a domain-tailored, multi-stage training framework and an iterative inference strategy, enhancing its capability to navigate the complexities of legal reasoning.
% Unilaw-R1 addresses these gaps through a domain-tailored
% Unilaw-R1 addresses these gaps through a domain-tailored

\section{Approach}

\subsection{Overview}

We propose a two-stage framework for legal reasoning model construction, as illustrated in Figure~\ref{2stage}. In the data generation stage, we construct a high-quality legal reasoning dataset, Unilaw-R1-Data, by leveraging a data distillation approach grounded in DeepSeek-R1 and incorporating an LLM-as-judge filtering mechanism \citep{xu2023mmbenchmark} to ensure annotation consistency and reasoning rigor. In the model training stage, we develop the Unilaw-R1 model based on Qwen2.5-7B-Instruct \citep{yang2024qwen2}, utilizing Supervised Fine-Tuning (SFT) in combination with the Group Relative Policy Optimization (GRPO) algorithm \citep{shao2024deepseekmath}. To further enhance reasoning performance, we introduce an iterative inference mechanism with a collaborative Assessor-Reviser agent setup, enabling the model to refine its reasoning trajectory for more accurate, coherent, and legally sound outputs. The overall process ensures that the model delivers structured, standardized outputs aligned with professional requirements.

\subsection{Data Construction}

\begin{figure*}[htbp] % figure 
        \centering % 
        \includegraphics[width=\linewidth]{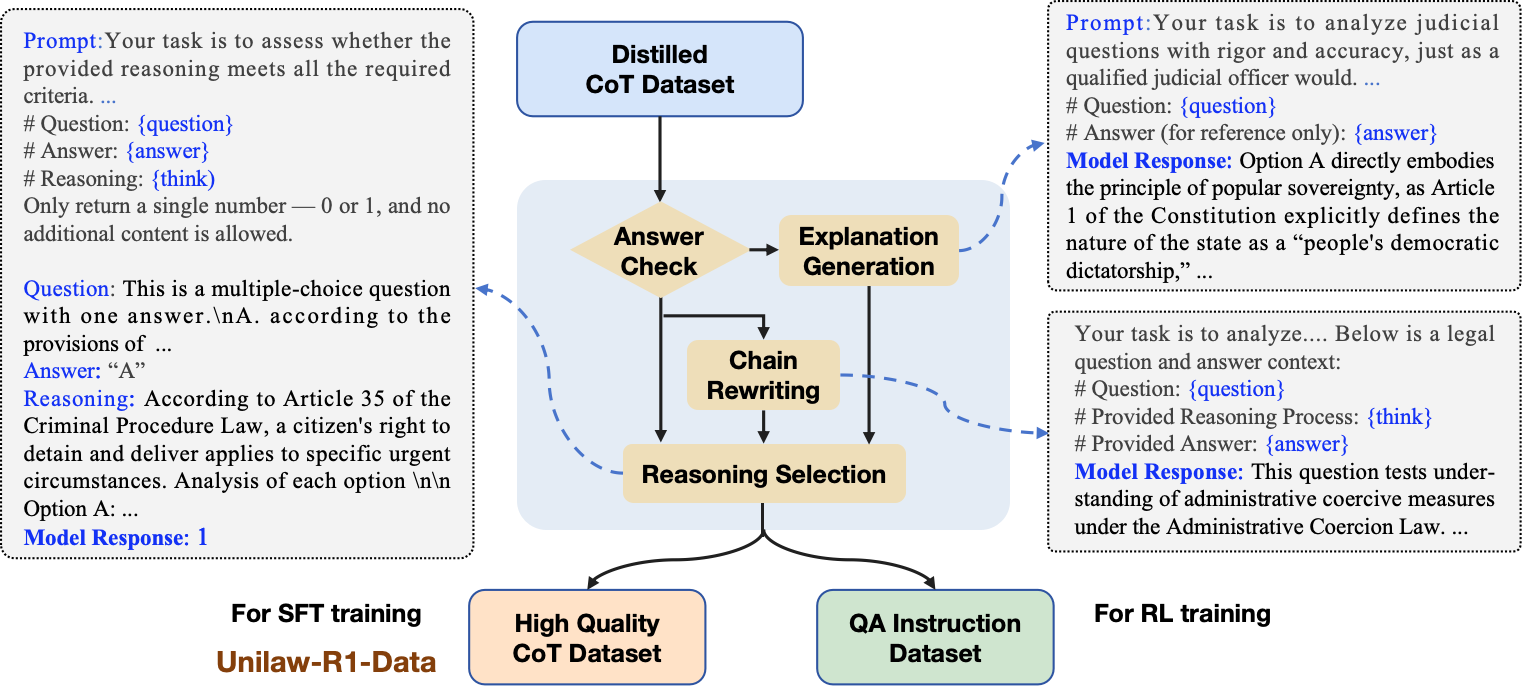}
        \caption{The pipeline of Data Construction (Stage 1): (1) Data Distillation, (2) Data Filtering, including Answer Check and Reasoning Selection, Chain Rewriting, and Explanation Generation. "Reasoning" represents the reasoning output, while "Model Response" refers to the evaluation process of the judgment model.} 
        \label{data_construct} 
    \end{figure*}

We aim to develop Unilaw-R1-Data, a high-quality  supervised fine-tuning dataset tailored for the legal domain. To this end, we designed a comprehensive data construction pipeline that filters and refines data for accuracy and reliability. We also rewrite samples to align with the syllogistic reasoning framework common in legal analysis. As shown in Figure \ref{data_construct}, the pipeline includes Answer Check, Chain Rewriting, Explanation Generation and Chain Selection, where an LLM evaluates DeepSeek-R1 outputs for correctness and scores the reasoning paths to ensure logical coherence.

\subsubsection{Data Source}
\label{sec:Data Source}

Unilaw-R1-Data consists of objective question answering entries in the legal domain, drawn from two primary sources: the open-source JEC-QA dataset \citep{yue2023disc} and proprietary data. JEC-QA includes 26, 365 multiple-choice questions, each with a question and four options. The proprietary portion includes 1, 700 multiple-choice question answering entries from National Judicial Examination of China from year 2015 to 2021. These were collected as PDFs, converted to markdown using Mineru \citep{wang2024mineruopensourcesolutionprecise}, and structured into question-answer pairs via regex-based extraction. All entries were manually reviewed for accuracy. From our proprietary data, 800 high-quality samples were retained to form the Unilaw-R1-Eval set for model evaluation.

\subsubsection{Data Processing}
Unilaw-R1-Data was constructed through a rigorous, multi-stage process involving data distillation and filtering. The dataset do not contain any answer explanations. To collect SFT examples, we first distilled multiple-choice questions into a question-thinking-answering format using the reasoning model DeepSeek-R1, following its official parameter configurations during distillation. 

Data filtering comprises four key components: \textbf{Answer Check}, \textbf{Chain Rewriting}, \textbf{Explanation Generation} and \textbf{Reasoning Selection}. In the Answer Check stage, we retain only those responses that strictly align with the reference answers. Specifically, any response generated by DeepSeek-R1 that diverges from the ground truth in the dataset is immediately excluded. We apply exact match to ensure correctness.

For the exactly matched responses, we sampling 10\% of it for Chain Rewriting. This component focuses on restructuring intermediate reasoning chains to ensure they conform to domain-specific logic and legal standards. For the unmatched responses, we sampling 10\% of it for Explanation Generation to keep the diverse style. We input both the question and corresponding answer into DeepSeek-V3 and ask it to output the explanation only. We integrate legal rules and definitions as rewriting and explanating constraints to ensure the reasoning paths remain consistent with normative legal interpretations. 

All generated chains from Chain Rewriting and Explanation Generation modules, along with those filtered by Answer Check module, are passed into the Reasoning Selection phase to evaluate the plausibility and legal soundness of multiple reasoning trajectories using the instruction model DeepSeek-V3 \citep{liu2024deepseek}. Responses are scored based on their adherence to legal reasoning principles, such as the correct application of rules, consistency with precedent, and logical coherence. These dimensions were employed to comprehensively evaluate the model's reasoning trajectory data. The model's reasoning path must not only lead to the correct answer but also demonstrate a valid and interpretable argumentative structure. When multiple valid paths exist, we prioritize those that align more closely with recognized legal standards and practices. 
% In the reasoning selection phase, we drew inspiration from the study by \cite{xie2024finnlp} and distilled seven key dimensions from it: internal consistency, term overlap rate, number of reasoning steps, logical coherence, content diversity, task-domain relevance, and alignment with task instructions. These dimensions were employed to comprehensively evaluate the model's reasoning trajectory data.
Further details on the experimental setup and findings are provided in Appendix \ref{sec:dataprompt}.

\subsubsection{Data Statistics}
After the data processing, we scored and filtered the reasoning paths, retaining only high-quality trajectories to construct the Unilaw-R1-Data for supervised fine-tuning, then randomly selected 8,000 QA entries — half from the unselected pool and half from Unilaw-R1-Data — for reinforcement learning. The Unilaw-R1-Data and Unilaw-R1-Eval datasets is presented in Table \ref{tab:data-statistic}. The table systematically details the descriptions of these datasets, including the data used stage, the question type, and average token length distribution of prompt, chain of thought reasoning and answer. 

The datasets include both knowledge-driven questions and case-based questions. Knowledge-driven questions assess the understanding of legal concepts, while case-based questions focus on the logical analysis of real-world legal scenarios. These two categories comprehensively cover a wide range of legal business scenarios. For evaluation, Unilaw-R1-Eval is categorized into knowledge and case-based subsets, and each question is also labeled with its specific legal domain, further details are provided in Appendix \ref{sec:statistics}.

% \begin{table}[htbp]
% \centering
% \small
% \begin{tabular}{@{}cccccc@{}}
% \toprule
% \multirow{2}{*}{Stage} & \multicolumn{2}{c}{Data Number} & \multicolumn{3}{c}{Token Length (avg.)} \\ \cmidrule(l){2-6} 
%                        & SC             & MC             & Prompt       & Think      & Answer      \\ \midrule
% SFT                    & 9534& 7001& 332& 723& 228\\ \midrule
% RL                     & 3320& 4780& 371& 488& 162\\ \bottomrule
% \end{tabular}
% \caption{The staticstics of Unilaw-R1-Data, including the data number of question type and average token length }
% \label{tab:data-statistic}
% \end{table}

\begin{table}[htbp]
\centering
\small
\begin{tabular}{@{}cccccc@{}}
\toprule
\multirow{2}{*}{\textbf{Stage}} & \multicolumn{2}{c}{\textbf{Data Number}} & \multicolumn{3}{c}{\textbf{Token Length}} \\ 
\cmidrule(l){2-3}  \cmidrule(l){4-6} 
                       & SC             & MC             & PRM       & THT      & ANS      \\ 
\midrule
Unilaw-R1-Data                    & 9534& 7001& 332& 723& 228\\ \midrule
Unilaw-R1-Eval                     & 426& 374& 176& -& 2\\ \bottomrule
\end{tabular}
\caption{The data statistics for Unilaw-R1-Data and Unilaw-R1-Eval, including the number of single-choice (SC) and multi-choice (MC) questions, as well as the average token lengths for prompts (PRM), chain-of-thought reasoning (THT), and answers (ANS).}
\label{tab:data-statistic}
\end{table}

\subsection{Training Method}
Unilaw-R1 is first trained via Supervised Fine-Tuning (SFT) using a high-quality legal reasoning dataset to enhance its reasoning ability. Building on this, we implement Group Relative Policy Optimization (GRPO) reinforcement learning to leverage legal question-answer data, incorporating a triple reward mechanism to improve both the accuracy of response formatting and content. The Stage 2 in Figure~\ref{2stage} intuitively summarizes the comprehensive training framework, illustrating the synergistic integration of the supervised learning and reinforcement learning components. Additional details about the training setup can be found in Appendix~\ref{sec:iter-exp}.

\subsubsection{Training Data Template}

%In this study, we employ Qwen2.5-7B-Instruct as the foundation model. The training data is structured as follows:  

This section outlines the data training format and its role in the subsequent training process.

\noindent
\textbf{SFT Training Data} \hspace{1em} During the Supervised Fine-Tuning (SFT) phase, each sample \( s \) in the training dataset \( S \) comprises three components, i.e., \( s = (x, c, y^*) \), where \( x \) denotes the question, \( c \) represents the reasoning trace formatted as \texttt{<think>...</think>}, and \( y^* \) corresponds to the answer, formatted as \texttt{<answer>...</answer>}.  
During the SFT stage, \( x \) is used as the input of the training set, \( c \) and \( y^* \) are used as the output of the training set.
This phase enables the model to learn structured legal reasoning patterns, refining its parameters to generate well-formed reasoning traces and accurate answers.  

\noindent
\textbf{RL Training Data} \hspace{1em} During the reinforcement learning (RL) phase, each sample \( s \) in the training dataset \( S \) consists of two components, i.e., \( s = (x, y^*) \), where \( x \) denotes the question and \( y^* \) represents the model’s output, which includes only the answer without reasoning traces.  
Reinforcement learning further enhances output quality by improving answer accuracy and ensuring compliance with the expected format.

\subsubsection{Supervised Fine-Tuning}
We initially performed Supervised Fine-Tuning on Qwen2.5-7B-Instruct using the LoRA efficient parameter tuning method to optimize key aspects of legal reasoning. The fine-tuning was conducted on the Unilaw-R1-Data dataset, incorporating a high-quality CoT reasoning process. This fine-tuning process effectively mitigated the reasoning failures observed when applying the general-purpose model to legal reasoning tasks. Following SFT, the model not only exhibited improved performance in legal reasoning but also learned to generate reasoning trajectories in the \texttt{<think>}...\texttt{</think>} format.

\subsubsection{Group Relative Policy Optimization}

During the reinforcement learning phase, we employ the Group Relative Policy Optimization (GRPO) algorithm. 
In each training iteration, \(G\) candidate outputs \(\{o_i\}_{i=1}^G\) are sampled from the old policy \(\pi_{\text{old}}\), each assigned a reward \(r_i\). The group-relative advantage \(A_i\) then computed as:
\begin{equation}
    A_i = \frac{r_i - \mu_{\{r\}}}{\sigma_{\{r\}}},
\end{equation}
where \(\mu_{\{r\}}\) and \(\sigma_{\{r\}}\) denote the mean and standard deviation of reward values within the group. Outputs exceeding group averages receive higher advantage values for prioritized optimization.
The policy update now maximizes the following objective function:

\begin{small}
\begin{equation}
\begin{aligned}
\mathlarger{\mathcal{J}_{\text{GRPO}}(\theta) }&\mathlarger{\mathlarger{= 
\mathbb{E}_{\mathbf{s} \sim P(\mathbf{S}),\, \{o_i\}_{i=1}^G \sim \pi_{\theta_{\text{old}}}(O|\mathbf{s})} }}\\
\frac{1}{G} \sum_{i=1}^G &\bigg \{\min\Big[ r_i A_i,\, \operatorname{clip}\left( r_i, 1 - \epsilon, 1 + \epsilon \right) A_i \Big] \\
&\quad - \beta\, D_{\text{KL}}\left( \pi_{\theta} \,\|\, \pi_{\text{ref}} \right) \bigg \},
\end{aligned}
\end{equation}
\end{small}
% \begin{equation}
% \small
% \begin{aligned}
% \mathcal{J}_{\text{GRPO}}(\theta) &= 
% \mathbb{E}_{\mathbf{s} \sim P(\mathbf{S}),\, \{o_i\}_{i=1}^G \sim \pi_{\theta_{\text{old}}}(O|\mathbf{s})} \\
% \frac{1}{G} \sum_{i=1}^G &\Bigg \{\min\Bigg[ r_i A_i,\, \operatorname{clip}\left( r_i, 1 - \epsilon, 1 + \epsilon \right) A_i \Bigg] \\
% &\quad - \beta\, D_{\text{KL}}\left( \pi_{\theta} \,\|\, \pi_{\text{ref}} \right) \Bigg \}, 
% \end{aligned}
% \end{equation}
\noindent
where \(r_i = \frac{\pi_{\theta}(o_i|\mathbf{v})}{\pi_{\theta_{\text{old}}}(o_i|\mathbf{v})}\) represents the importance sampling ratio that quantifies the relative likelihood of generating output \(o_i\) under the new policy \(\pi_{\theta}\) compared to the old policy \(\pi_{\theta_{\text{old}}}\); \(A_i\) denotes the group-relative advantage, calculated by normalizing each reward with respect to the group’s mean and standard deviation to emphasize outputs that surpass the group average; the clipping operator \(\text{clip}\!\left( \cdot \right)\) restricts the update magnitude within the trust region \([1 - \epsilon, 1 + \epsilon]\) to avoid destabilizing large parameter changes; the minimum operation between the unclipped term \(r_i A_i\) and its clipped counterpart ensures a conservative update that balances aggressive improvements with training stability; and finally, $D_{\text{KL}}\!\left(\pi_{\theta} \,\|\, \pi_{\text{ref}}\right)$ is the KL divergence and $\beta$ is the hyper-parameter.

\subsubsection{Reward Function Design}
In the process of training the reward model based on GRPO, we employs three reward mechanisms: accuracy reward, format reward and legal validity reward.

\noindent
%qwen72b
\textbf{Accuracy Reward}\hspace{1em}We use the rule-based regular expressions methods to extract the content within the \texttt{<answer>}...\texttt{</answer>} tags from the model’s output.  This extracted answer is then compared against a reference solution. If the output within the \texttt{<answer>} tags is semantically consistent with the reference answer, a reward score of 1 is assigned; otherwise, it receives a score of 0. The accuracy reward function is defined as follows:
\begin{equation}
\begin{aligned}
R_{\text{Acc}}(y, y^*) = 
\begin{cases}
1, & \text{if } y=y^* \\
0, & \text{otherwise}
\end{cases}
\end{aligned}
\end{equation}
where $y$ is model's output (from \texttt{<answer>}... \texttt{</answer>} tags). $y^*$ is the standard answer.

\noindent
\textbf{Format Reward}\hspace{1em}We encourage outputs that include a sequence of reasoning steps enclosed within \texttt{<think>...\texttt{</think>}} tags and a concise final answer enclosed within \texttt{<answer>...\texttt{</answer>}} tags. A format incentive score of 1 is awarded if all four tags appear exactly once with no extraneous content outside these tags; otherwise, a score of 0 is assigned. The format reward function is defined as follows:
\begin{equation}
    R_{\text{Fmt}}(y) = 
\begin{cases}
1, & \text{if the format matches} \\
0, & \text{otherwise}
\end{cases}
\end{equation}
where \(y\) denotes the model's output. Format matching indicates that the output strictly adheres to the specified format by containing exactly one pair of \texttt{<think>} tags and one pair of \texttt{<answer>} tags, with no additional content outside these tags.
\\

%judge model
\noindent
\textbf{Legal Validity Reward}\hspace{1em}The precise and contextually accurate answers are essential in legal scenarios. To ensure this, we employ an instruct model to evaluate whether the reasoning model’s output aligns with the intended legal solution. This approach offers a more robust assessment compared to traditional rule-based methods. The model-based verifier plays a crucial role in ensuring the correctness of responses, particularly in complex and nuanced legal contexts. 

The evaluation criteria in the prompt provided to the LLM are largely aligned with those used in the chain rewriting task, and more details are outlined in Appendix~\ref{sec:Prompt of Judging}. These instructions include the application of key legal principles, such as \textbf{syllogism}, which follows a structure of a major premise (general legal rule), a minor premise (specific case fact), and a conclusion (legal inference). Additionally, the model is required to adhere to formal legal citation standards. Based on the model’s output and its adherence to the rules specified in the prompt,  the output score can be one of the following values: 2, 1, or 0. The score is determined by the extent to which the model’s output aligns with the expected answer. The legal validity reward function is thus defined as follows:

\begin{small}
\begin{equation}
R_{\text{Legal}}(y, y^*) = 
\begin{cases}
2, & \text{if }  y \ \text{consistent with} \  y^* \\
1, & \text{if }  y \ \text{partially consistent with} \   y^* \\
0, & \text{otherwise} 
\end{cases}
\end{equation}
\end{small}
% \begin{equation}
% \scalebox{0.9}{$
% R_{\text{Legal}}(y, y^*) = 
% \begin{cases}
% 2, & \text{if }  y \ \text{consistent with} \  y^* \\
% 1, & \text{if }  y \ \text{partially consistent with} \   y^* \\
% 0, & \text{otherwise}
% \end{cases}
% $}
% \end{equation}

\noindent where \( y \) represents the model's output (extracted from the \texttt{<think>}... \texttt{</think>} tags), and \( y^* \) is the standard legal answer.

\noindent
\textbf{Total Reward}\hspace{1em}The total reward is computed as the weighted sum of the above rewards, formulated as follows:
\begin{equation}
    \mathcal{R} = \alpha R_\text{Acc} + \beta R_\text{Fmt} + \gamma R_\text{Legal},
\end{equation}
where $\alpha = 0.9$, $\beta = 0.1 $, and $ \gamma = 0.1$.

\subsection{Iterative Inference}

\begin{figure*}[htbp] % figure 
    \centering % 
    \includegraphics[width=0.9\textwidth]{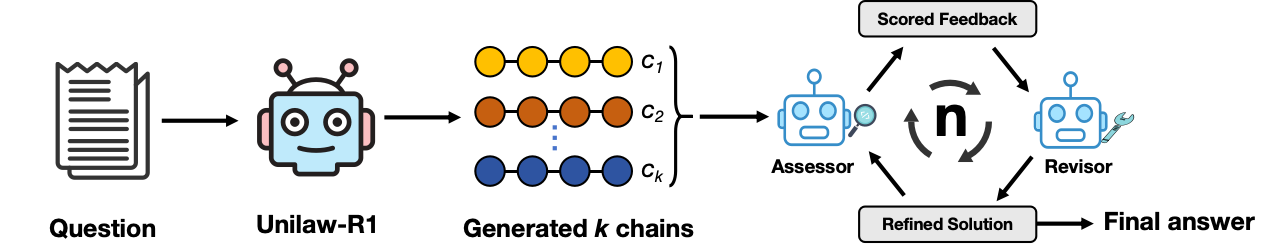}
    \caption{Iterative inference pipeline, consisting of four main stages: sampling, reviewing, refinement, and final answer selection. The reviewing and refinement stages involve a multi-agent setup, with separate Assessor and Revisor agents. } % 
    \label{inference} % 
\end{figure*}

To enhance response quality in legal language generation, we propose an \textit{iterative inference} framework, as shown in Figure \ref{inference}. It consists of four main stages: sampling, reviewing, refinement, and final answer selection. The reviewing and refinement stages involve a multi-agent setup, with separate Assessor and Revisor agents. These two stages are applied over \( n \) iterations to progressively refine candidate responses.

\subsubsection{Sampling Chains}

Given an input prompt \( x \), we first generate a set of \( k \) diverse candidate responses using the post-trained legal reasoning language model \( \mathcal{M}_{\text{Unilaw}} \). These candidates are generated by sampling with different parameters to ensure diversity among the outputs:
\begin{equation}
\{ y_i^{(0)} \}_{i=1}^k \sim \text{Sampling}(\mathcal{M}_{\text{Unilaw}}(x), k)
\end{equation}
Here, \( y_i^{(0)} \) denotes the \( i \)-th candidate in the initial generation batch ($Iter = 0$).

\subsubsection{Assessing Candidate Responses}

Each candidate \( y_i^{(t)} \) at iteration \( t \) is evaluated using a Assessor agent \( \mathcal{K} \), which produces a step-wise quality score and an actionable feedback:
\begin{equation}
fb_i^{(t)} = \mathcal{K}(x, y_i^{(t)})
\end{equation}

The agent takes a chain as input, scores each step, and then identifies problematic steps based on these scores. Responses that fall below a predefined threshold score are flagged for refinement, with potential solutions for improvement in the next stage. A one-shot prompt is provided to guide the reviewer on how to score each step in the chain and generate targeted feedback. The prompt for the Assessor can be found in Appendix \ref{sec:iterativeprompt}.

\subsubsection{Revising Problematic Responses}

A Revisor agent \( \mathcal{F} \) is then applied to the selected low-quality responses to improve their relevance, coherence, or correctness. For each low-scoring candidate:
\begin{equation}
y_i^{(t+1)} = 
\begin{cases}
\mathcal{F}(x, fb_i^{t}, y_i^{(t)}), & \text{if } s_i^{(t)} < \tau \\
y_i^{(t)}, & \text{otherwise}
\end{cases}
\end{equation}

By highlighting specific errors in the reasoning chain, the targeted feedback enables the Revisor to address mistakes more effectively, as it clearly identifies which step are incorrect; We use 1-shot prompt to teach the Revisor how to fix the error and improve a reasoning chain based on targeted feedback. The refined candidate are then passed into the next review iteration. The prompt for the Reviser is shown in Appendix \ref{sec:iterativeprompt}.

\subsubsection{Final Answer Selection}

This review-refine loop continues for \( n \) total iterations. At the end of each iteration, we evaluate whether the refined solutions represent an improvement useing the outcome reward model (ORM). Specifically, we compare the 2$k$ reasoning chains - $k$ initial and $k$ refined - and retain the top $k$ based on their global ORM scores. After the final iteration \( t = n \), the final answer is selected via self-consistency over the retained top $k$ chains.
This iterative inference process effectively combines generation diversity with feedback-driven refinement to produce high-quality legal responses.

\section{Experiment}
\label{s3}

\subsection{Datasets}
We evaluate our model on Unilaw-R1-Eval dataset and two additional Chinese legal domain multi-task benchmarks: LawBench \citep{fei2024lawbench} dataset and LexEval \citep{li2024lexeval}. LawBench assesses the legal capabilities of LLMs across three cognitive levels: memory, understanding, and application. It comprises 20 tasks with various formats, including multiple-choice, extraction, generation, and regression, simulating real-world legal scenarios such as statute prediction, case analysis, and legal consultation. LexEval, the largest and most comprehensive Chinese legal benchmarking dataset, evaluates performance of LLMs across six cognitive abilities defined by the LexCog taxonomy: memory, understanding, logical reasoning, discrimination, generation, and ethics. It consists of 14,150 entries across 23 legal tasks, providing a diverse set for evaluating LLM performance.

We evaluate our model in zero-shot settings. The inputs to the LLMs are only instructions and queries. We use Accuracy and F1 to to evaluate the Unilaw-R1-Eval data. For LawBench and LexEval datasets, we employ automated evaluation methods tailored to the diverse task types within their benchmarks, ensuring objective and consistent assessment of large language models in legal contexts.

\begin{table*}[!t]
\label{tab:model_compare}
% \vspace{10pt}
\resizebox{0.95\linewidth}{!}{% 
    \centering
    \renewcommand{\arraystretch}{1.1} % 
    \begin{tabular}{>{\large}l >{\large}c *{5}{>{\large}c} >{\large}c} 
    \toprule[2pt]
    \textbf{Model} & \textbf{Parameters} & \textbf{LawBench} & \textbf{LexEval} & \textbf{Unilaw-R1-Eval} & \textbf{Avg.}(\%) \\
    \midrule[1pt]
    DeepSeek-R1 & 671B & 61.8 & 67.2 & 55.2 & 61.4 \\
    DeepSeek-V3 & 671B & 61.3 & 65.7 & 50.6 & 59.2 \\
    DeepSeek-R1-Distill-Qwen-32B & 32B & 57.0 & 65.2 & 42.6 & 54.9 \\
    Qwen-2.5-32B-Instruct & 32B & 63.8 & 66.9 & 42.2 & 57.6 \\
    DeepSeek-R1-Distill-Qwen-14B & 14B & 51.8 & 54.8 & 24.0 & 43.5 \\
    Qwen-2.5-14B-Instruct & 14B & 58.3 & 64.3 & 29.4 & 50.6 \\
    DeepSeek-R1-Distill-Qwen-7B & 7B & 38.3 & 47.3 & 23.6 & 36.4 \\
    Qwen-2.5-7B-Instruct & 7B & 52.3 & 57.8 & 29.9 & 46.6 \\
    \midrule[1pt]
    Unilaw-R1-SFT & 7B & 52.2 & 58.6 & 33.3 & 48.0 \\
    Unilaw-R1-RL & 7B & 54.2 & 60.6 & 35.6 & 50.4 \\
    \textbf{Unilaw-R1} & 7B & 56.6 & 63.5 & 39.5 & 53.2 \\
    
    \bottomrule[2pt]
    \end{tabular}
}
\centering
\caption{Accuracy evaluation of Unilaw-R1-SFT and Unilaw-R1 on different legal benchmarks.}
\label{mainres}
\vspace{-5pt}
\end{table*}

\subsection{Baselines}
To comprehensively evaluate the reasoning capabilities of Unilaw-R1 in legal scenarios, we conducted a thorough comparative assessment against multiple baseline models. These models include DeepSeek-R1, DeepSeek-V3, DeepSeek-R1-Distill-Qwen-32B, DeepSeek-R1-Distill-Qwen-14B, DeepSeek-R1-Distill-Qwen-7B, Qwen-2.5-32B-Instruct, Qwen-2.5-14B-Instruct, Qwen-2.5-7B-Instruct, Unilaw-R1-SFT and Unilaw-R1-RL. The selection of these models encompasses a spectrum ranging from lightweight to high-performance architectures, taking into account factors such as reasoning capability and computational resource consumption. This comprehensive comparison aims to provide a holistic evaluation the performance of Unilaw-R1 within legal applications.
\subsection{Main Results}
Table~\ref{mainres} presents the results of our comprehensive benchmarking evaluation across multiple legal business scenarios. Unilaw-R1 demonstrated notable performance advantages despite its lightweight 7B parameter size. Leveraging a two-stage training framework, it achieved an average score of 53.2\%. Remarkably, Unilaw-R1 outperformed all participating models of similar scale and even achieved performance comparable to the much larger DeepSeek-R1-Distill-Qwen-32B (54.9\%). 
Following domain-specific training, Unilaw-R1 exhibited significant performance improvements in other legal benchmarks such as LawBench, LexEval, surpassing Qwen-2.5-7B-Instruct (46.6\%) by an average margin of 6.6\%.

Fine-tuning Qwen-2.5-7B-Instruct on Unilaw-R1-Data and RL data resulted in the Unilaw-R1-SFT and Unilaw-R1-RL models, with average performance improvements of 1.4\% and 3.8\%，respectively. 
These results demonstrate strong cross-task generalization and effectiveness in legal applications.

\begin{table}[htpb]
% \vspace{10pt}
\centering
\resizebox{1.0\linewidth}{!}{%
\renewcommand{\arraystretch}{1.1} % 
\begin{tabular}{@{}lcccc@{}}
\toprule[2pt]
\multirow{2}{*}{\textbf{Method}} & \multicolumn{1}{c}{\textbf{SC}} & \multicolumn{2}{c}{\textbf{MC}} & \multicolumn{1}{c}{\textbf{Avg.}} \\
\cmidrule(r){2-2} \cmidrule(lr){3-4} \cmidrule(l){5-5}
    & Acc.(\%)                   & Acc.(\%)         & F1    & Acc.(\%)  \\
\midrule[1pt]
Zero-shot CoT               & 53.8          & 23.2         & 67.4   & 39.5  \\
Best-of-$k$ ($k=10$)            & 62.3          & 25.7         & 67.9    & 45.2 \\
Majority Vote               & 56.8          & 33.1         & 66.6   & 45.7  \\
\midrule[1pt]
Iterative Infer. ($Iter=1$)   & 65.5          & 33.4         & 71.9    & 50.5 \\
Iterative Infer. ($Iter=2$) & \textbf{66.3}          & 33.8     & \textbf{72.2}  & 50.6  \\
Iterative Infer. ($Iter=3$)   & 65.7    & \textbf{34.3}   & 71.3  & \textbf{51.0}   \\
\bottomrule[2pt]
\end{tabular}
}
\caption{Performance comparison of Unilaw-R1 with different inference methods on the Unilaw-R1-Eval benchmark.}
\label{absres}
\vspace{-10pt} %
\end{table}

\subsection{Ablation Study}
We conducted an ablation study to assess the performance impact of different inference strategies for the Unilaw-R1 model, as well as the convergence behavior of various combinations of reinforcement learning reward functions for the Unilaw-R1-SFT model, using the Unilaw-R1-Eval benchmark. 

As shown in Table \ref{absres}, we compared zero-shot CoT \citep{wei2022chain}, best-of-$k$ sampling, majority vote \citep{wangself} and iterative inference methods across single-choice (SC) and multi-choice (MC) tasks. The zero-shot CoT baseline achieved 53.8\% accuracy on SC tasks and 23.2\% on MC tasks. Implementing best-of-$k$ ($k=10$) sampling and majority vote led to improvements, raising the average accuracy from 39.5\% to 45.2\% and 45.7\%, respectively.
The iterative inference approach demonstrated more substantial gains. With a single iteration, SC accuracy increased to 65.5\% and MC to 33.4\%. The performance gains form further iterations were limited: the second iteration achieved 66.3\% accuracy on SC and 33.8\% on MC, while the third iteration reached 65.7\% (SC) and 34.3\% (MC), respectively.
\begin{figure}[htpb]
\vspace{-10pt}
\centering
\includegraphics[width=0.9\linewidth]{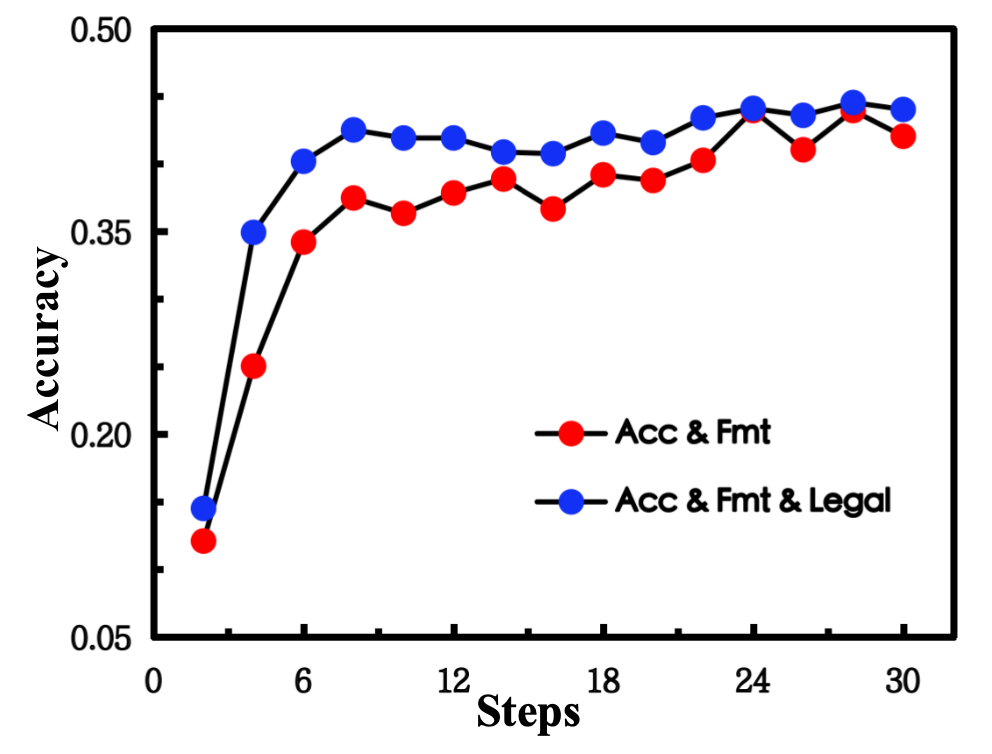}
\caption{Comparison of convergence behavior of Unilaw-R1-SFT under different combinations of reinforcement learning reward functions on the Unilaw-R1-Eval benchmark.}
\label{convergence}
\vspace{-10pt}
\end{figure} 
These results indicate that iterative inference significantly enhances model performance, particularly in the first iteration. However, additional iterations offer marginal improvements, suggesting a trade-off between computational cost and performance gains. Therefore, a single iteration of refinement may provide an optimal balance for practical applications.

As shown in Figure \ref{convergence}, we also compared two variants of Unilaw-R1-SFT: one with accuracy and format rewards (Acc \& Fmt), and one with an additional legal reward (Acc \& Fmt \& Legal). The latter showed faster convergence and higher accuracy, highlighting the effectiveness of the legal reward function.

\section{Conclusion}
\label{s4}
We introduce Unilaw-R1, a legal-domain reasoning LLM that combines distilled chain-of-thought data, a two-stage Supervised Fine-Tuning (SFT) followed by Reinforcement Learning (RL) training pipeline, and iterative inference multi-agent setup. This approach addresses data fragmentation, opaque reasoning, and poor generalization, achieving strong performance on legal benchmarks. Additionally, we propose a legal benchmark Unilaw-R1-Eval, which plays a critical role in assessing the model's performance in real-world legal scenarios.

\section*{Limitations}

\label{Limitations}
Despite notable advancements, our model faces several limitations:

\noindent
\textbf{Limited Training Data Coverage}: Currently, training data is confined to objective legal multi-choice questions, and it has not yet reached the satisfactory target. Future training will be expanded to a broader range of legal datasets. 

\noindent
\textbf{Single-Modality Architecture}: The model text-only architecture hinders its ability to process legal documents containing visual elements such as charts and tables. We plan to consider multimodal extension to address this limitation. 

\noindent
\textbf{Insufficient Evaluation of CoT Reasoning:} Our current evaluation compares model outputs against referenced answers but lacks analysis of the model's step-by-step legal reasoning. Future evaluations will focus on assessing the model's ability to perform structured legal reasoning, such as syllogistic reasoning, to align with legal standards.

We are committed to addressing the aforementioned limitations, expanding our model's application to emerging domains, and promoting broader adoption to strengthen legal risk management and compliance, ultimately increasing real-world impact and applicability.

% \section*{Acknowledgments}

% This document has been adapted
% by Steven Bethard, Ryan Cotterell and Rui Yan
% from the instructions for earlier ACL and NAACL proceedings.

% Bibliography entries for the entire Anthology, followed by custom entries
%\bibliography{anthology,custom}
% Custom bibliography entries only
\bibliography{main}

\clearpage
\appendix

\section{Prompts of Data Construction}
\label{sec:dataprompt}
Throughout the data construction pipeline, we designed prompts tailored to four critical stages: data distillation, explanation generation, chain-of-thought rewriting, and reasoning selection. These prompts were carefully crafted to guide the model in producing high-quality, logically consistent, and legally grounded outputs at each stage.

\subsection{The prompt of data distillation}
\label{dsprompt}

\begin{figure}[h]
\centering
\includegraphics[width=0.49\textwidth]{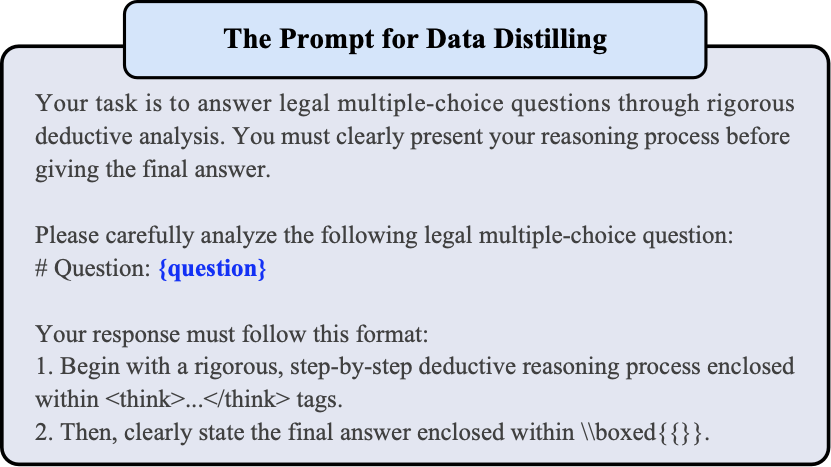}
\caption{The prompt of data distillation that we used for DeepSeek-R1.}
\label{fig:data-distill}
\end{figure}

In the data distillation phase, we drew inspiration from the official prompt design of DeepSeek-R1 and adapted it to the legal domain. Our prompt, illustrated in Figure~\ref{fig:data-distill}, was designed to elicit clear, structured reasoning traces from the base model. It ensured that the distilled responses were both informative and aligned with the expected chain-of-thought (CoT) format, serving as foundational supervision data for subsequent training stages.

\subsection{The prompt of explanation generation}
\label{The Prompt of explanation generation}

\begin{figure}[h]
\centering
\includegraphics[width=0.49\textwidth]{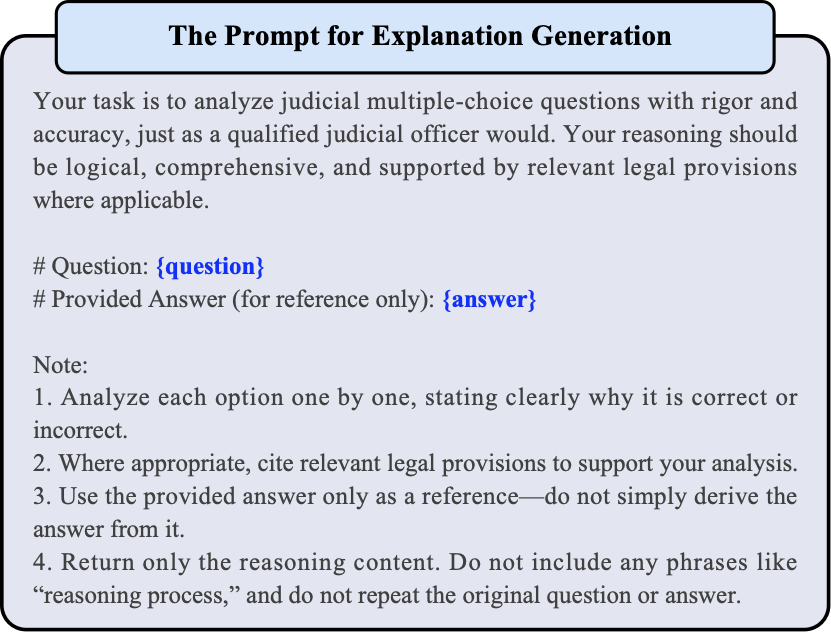}
\caption{The prompt of explanation generation that we used for DeepSeek-V3.}
\label{fig:data-eg}
\end{figure}

During the initial stage of data screening, we applied a regex-based answer check to filter the responses. For those that failed this check, we utilized the instruction-following model DeepSeek-V3 to regenerate explanations, providing it with the original question and answer as context. The specific prompting strategy used for explanation generation is illustrated in Figure \ref{fig:data-eg}.

\subsection{The prompt of chain rewriting}
\label{The Prompt of chain rewriting}
 
\begin{figure*}[h!] % figure 
    \centering % 
    \includegraphics[width=0.98\textwidth]{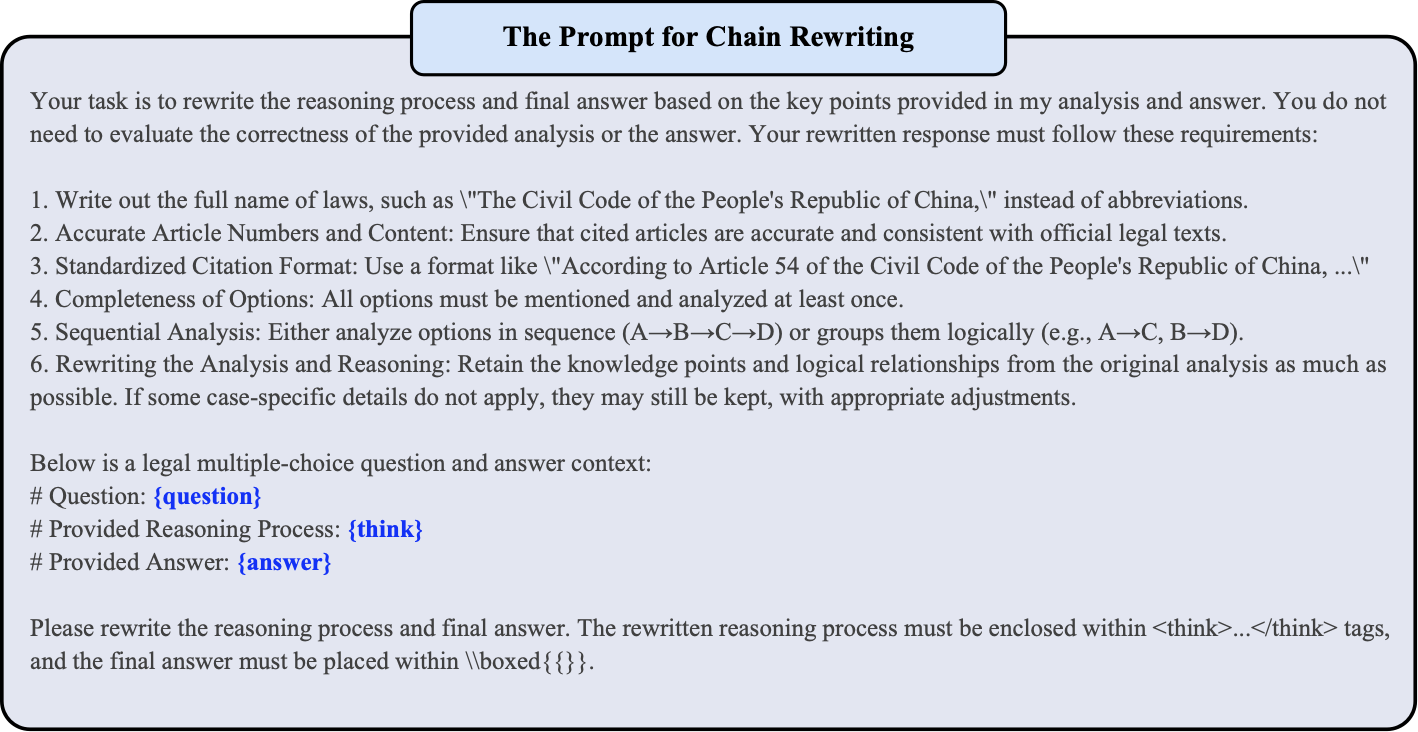} 
    \caption{The prompt of chain rewriting that we used for DeepSeek-V3.} % 
    \label{fig:chain_rewrite_prompt} % 
\end{figure*}

To preserve reasoning diversity, we randomly sampled 10\% of the examples that passed the answer check stage for chain rewriting. These samples were then used to generate alternative reasoning chains by leveraging the instruction-following capabilities of the DeepSeek-V3 model. Specifically, we provided the model with the original question, reference answer, and existing reasoning as context, prompting it to reconstruct the reasoning process. This approach introduces variation in logical pathways while maintaining answer consistency. The detailed prompting strategy used for this reasoning chain rewriting is illustrated in Figure \ref{fig:chain_rewrite_prompt}.

\subsection{The prompt of reasoning selection}
\label{The Prompt of reasoning selection}

\begin{figure*}[h!] % figure 
    \centering % 
    \includegraphics[width=0.98\textwidth]{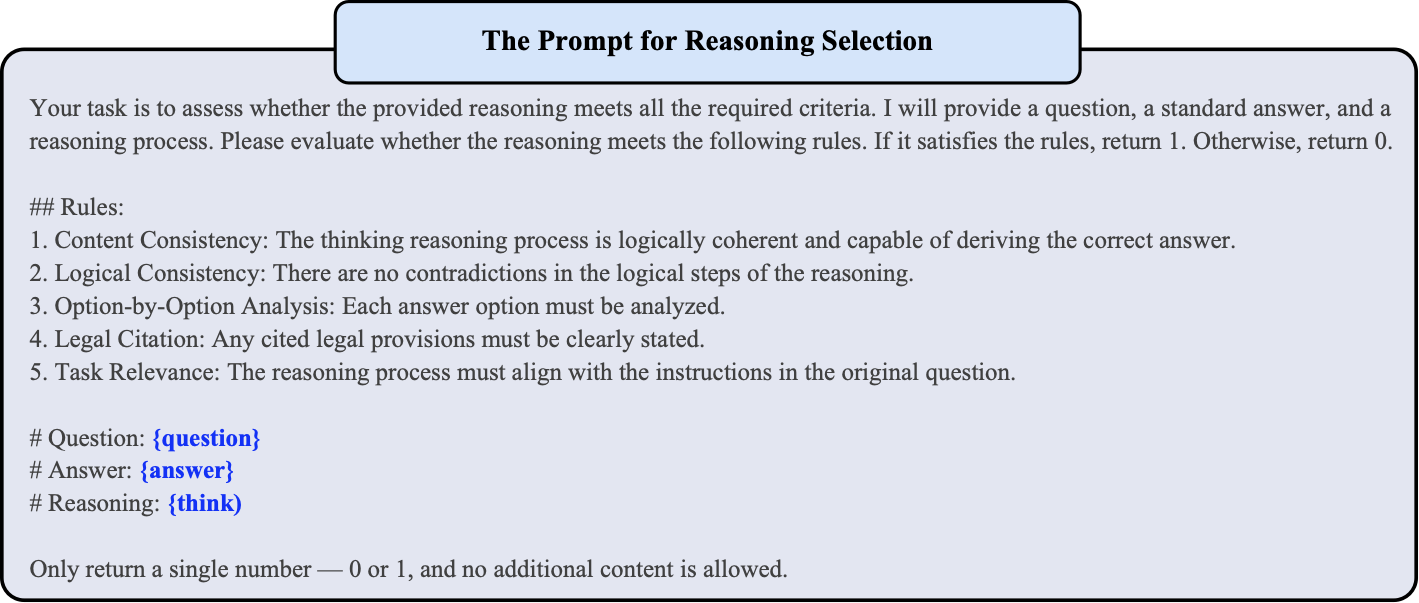} 
    \caption{The prompt for reasoning selection that we used for DeepSeek-V3.} % 
    \label{fig:data-rs} % 
\end{figure*}

Finally, to ensure the generation of high-quality reasoning trajectories, we introduced a reasoning selection data screening process. In this stage, we proposed five specific evaluation criteria to assess the model's reasoning performance. These criteria were carefully crafted to align with the core elements of effective legal reasoning. Furthermore, we designed and refined the prompt shown in Figure ~\ref{fig:data-rs} to guide the model toward generating accurate and interpretable responses.

In the initial preprocessing step, we conducted a detailed evaluation of the model-generated reasoning using the DeepSeek-V3 instruction model. This evaluation followed five predefined judgment criteria. For each reasoning output, a binary score of 1 was assigned if it met the criterion, and 0 otherwise. This binary scoring scheme (0/1) was applied systematically to ensure the consistency, reliability, and stability of the evaluation process.

\subsection{The statistics of Unilaw-R1-Eval}
\label{sec:statistics}

The Unilaw-R1-Eval comprises 800 curated comparative question-answer pairs, and we further constructed in a fine-grained and domain-relevant manner. These samples are categorized to reflect the diverse challenges encountered in real-world legal reasoning. More detailed statistics of question types are summarized in Table \ref{tab:ueval-statistic}.

We provide a categorical statistical analysis of the dataset through two concentric pie charts. Each chart corresponds to one of the two question formats included in the benchmark: single-choice (SC) and multi-choice (MC).

\begin{figure}[!t]
    \centering
    \begin{minipage}[t]{0.48\textwidth}
        \centering
        \includegraphics[width=\textwidth]{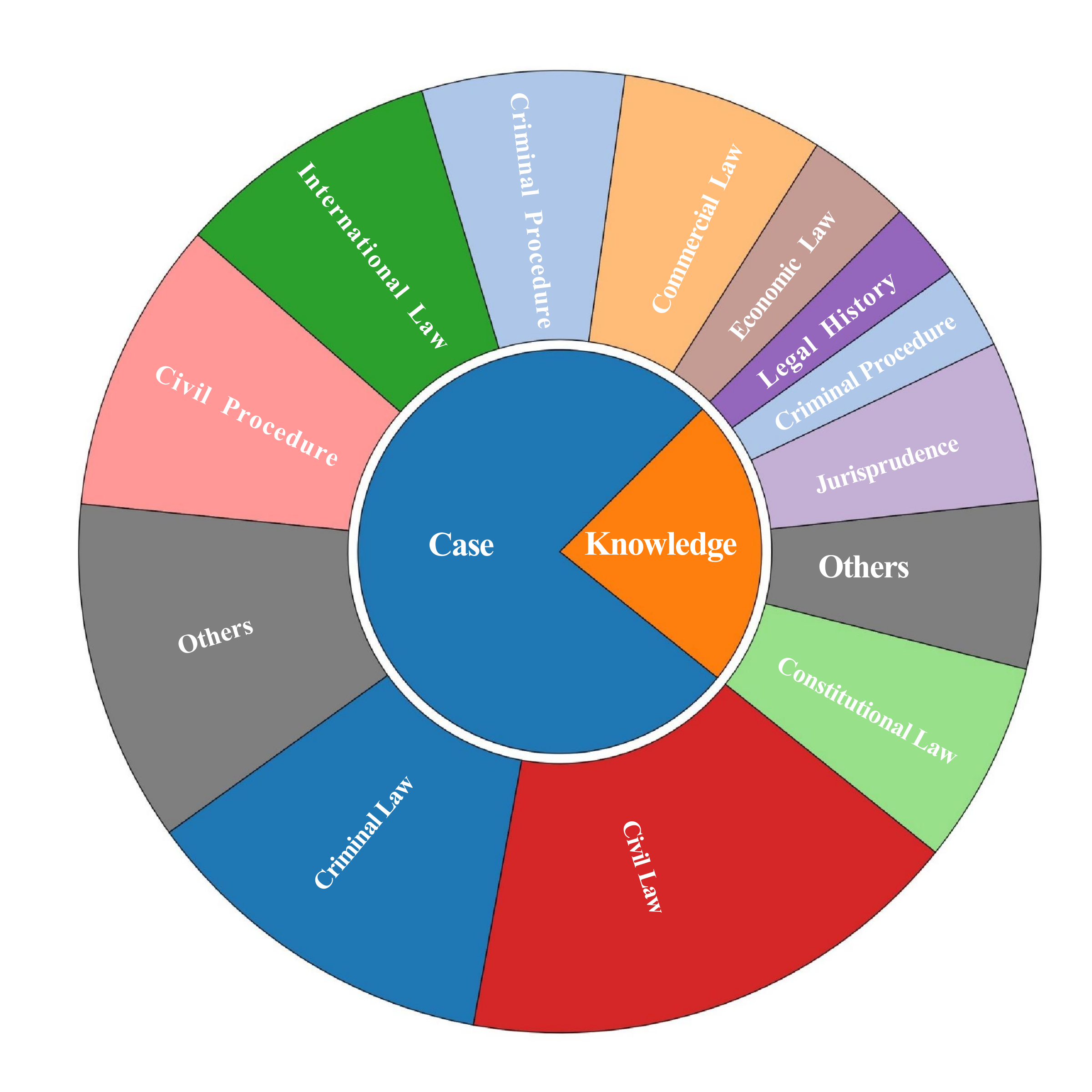}
        \subcaption{Single-choice question distribution in Unilaw-R1-Eval.}
        \label{sceval}
    \end{minipage}
    \hfill
    \begin{minipage}[t]{0.48\textwidth}
        \centering
        \includegraphics[width=\textwidth]{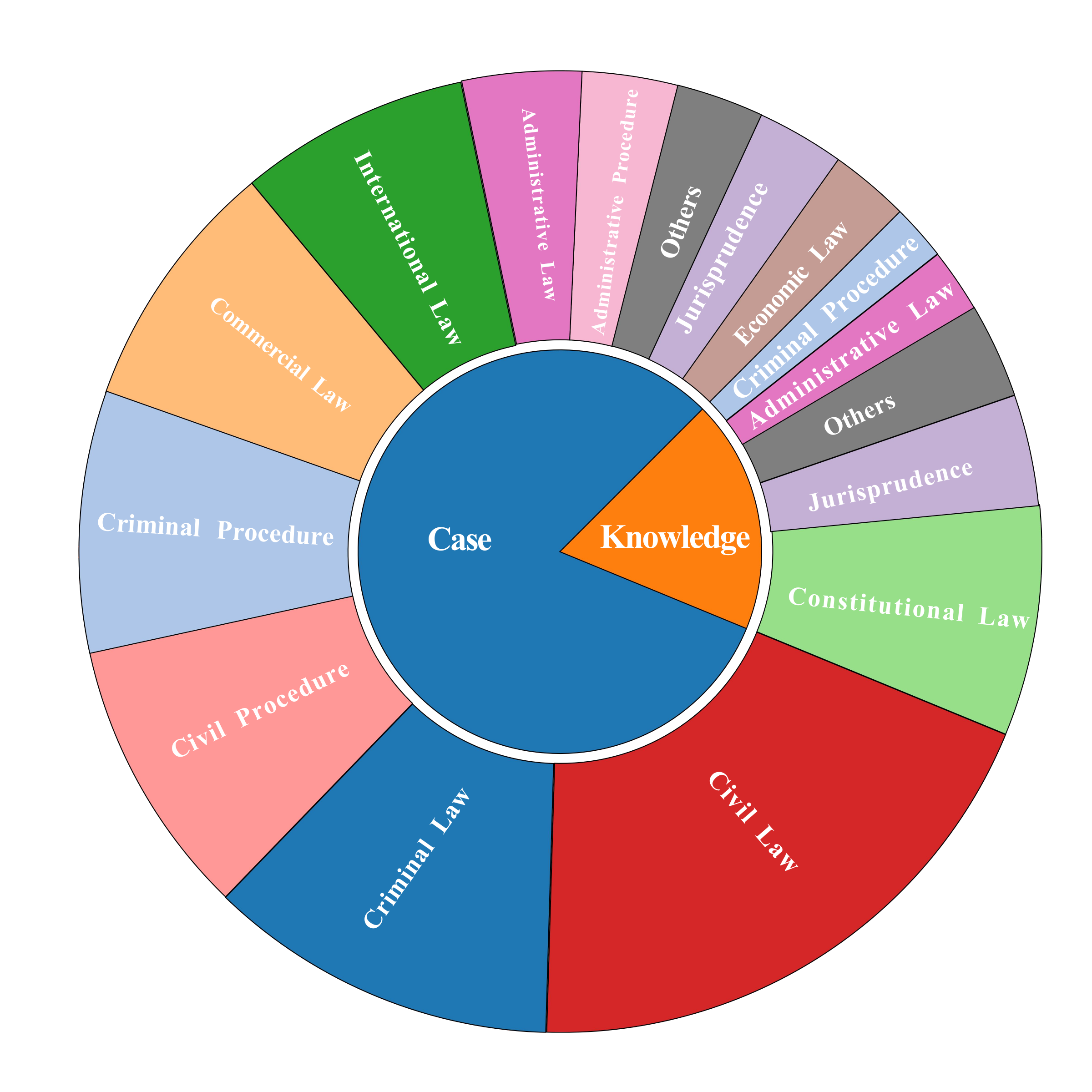}
        \subcaption{Multi-choice question distribution in Unilaw-R1-Eval.}
        \label{mceval}
    \end{minipage}
    \caption{Distribution of question types and legal subdomains in Unilaw-R1-Eval. The figure presents categorical statistics for both single-choice (SC) and multi-choice (MC) legal questions. The inner rings distinguish between knowledge-driven and case-driven types, while the outer rings represent their distribution across legal subdomains.}
\label{fig:unilaw-r1-data}
\end{figure}

As illustrated in Figure \ref{fig:unilaw-r1-data}(a), the chart visualizes the distribution of question types for the single-choice tasks, divided into two main categories:
\begin{itemize}
    \item \textbf{Case-driven} questions, which focus on logical reasoning and judgment over real or hypothetical legal scenarios.
    \item \textbf{Knowledge-driven} questions, which test the model's mastery of legal definitions, statutes, and normative concepts.
\end{itemize}

These above two categories represent complementary dimensions of legal AI: foundational legal knowledge and applied legal reasoning. Together, they cover a broad spectrum of legal domains allowing for domain-specific performance insights, as the outer ring shows. The legal subdomains include \textit{"Criminal Law", "Criminal Procedure", "Labor Law", "Commercial Law", "International Law", "Constitutional Law", "Civil Law", "Civil Procedure", "Legal History", "Jurisprudence", "Intellectual Property",  "Economic Law", "Administrative Law"}, and \textit{"Administrative Procedure"}. This layered categorization enables granular evaluation of a model’s capabilities in both conceptual understanding and real-world legal problem-solving.

As shown in Figure \ref{fig:unilaw-r1-data}(b), the chart reflects the distribution of multi-choice questions, which require models to evaluate multiple legal options simultaneously. These tasks often demand more comprehensive reasoning chains and sensitivity to nuanced distinctions between legal provisions.
Similar to the single-choice chart, the inner ring categorizes questions into knowledge-driven and case-driven types, while the outer ring provides a domain-level breakdown. The multi-choice questions particularly emphasize complex decision-making scenarios, such as those involving overlapping legal principles or multiple liable parties.

By providing detailed categorization for both question types and domain coverage, Unilaw-R1-Eval offers a rigorous, fine-grained benchmark for assessing legal-domain LLMs across knowledge comprehension, reasoning reliability, and generalization capacity. This dual-structured evaluation framework is instrumental for identifying both model strengths and performance bottlenecks across varied legal tasks.

\begin{table}[htpb]
\vspace{10pt}
\centering
\begin{tabular}{@{}lccc@{}}
\toprule
& \textbf{Knowledge} & \textbf{Case} & Total \\
\midrule[1pt]
Single-Choice          & 99          & 327         & 426  \\
Multi-Choice           & 70          & 304         & 374   \\
\midrule
All            & 169          & 631        & 800    \\
\bottomrule
\end{tabular}
\caption{The statistics of question types in Unilaw-R1-Eval.}
\label{tab:ueval-statistic}
\end{table}

\section{Prompt of Legal Validity Reward}
\label{sec:Prompt of Judging}

To enhance the alignment of the model’s outputs with legal correctness during reinforcement learning, we incorporate a model-based feedback mechanism. Specifically, we utilize an instruction language model Qwen2.5-7B-Instruct as a verifier to assess the quality of the reasoning trajectories generated by the policy model. This verifier evaluates each response against predefined legal reasoning criteria, including logical consistency, legal validity, and alignment with the expected legal outcome.

The model-based feedback is then used as a reward signal in the RL fine-tuning stage, replacing or complementing traditional rule-based or reference-based reward designs. This strategy enables the training process to dynamically adjust based on nuanced legal judgments rather than relying solely on static ground-truth answers. By leveraging the LLM’s own legal reasoning capabilities, we introduce a more flexible and context-aware reinforcement signal that supports the development of high-quality, legally sound responses.

The evaluation criteria used in the Legal Validity prompt are largely consistent with those in chain-of-thought rewriting, with an added emphasis on syllogistic reasoning in legal contexts, applying legal rules to case facts to derive conclusions.
\begin{itemize}
\item \textbf{Choice Analysis:} This emphasizes completeness by systematically analyzing each option either sequentially or in groups, ensuring that all answer choices are explicitly considered. Inaccurate or incomplete analysis may indicate failures in this structured deductive reasoning process, particularly when syllogistic reasoning is required.
\item \textbf{Legal Format:} This assesses the accuracy and consistency of cited legal provisions, including article numbers and their content, which should align with official legal texts. Additionally, it requires writing out the full names of laws rather than abbreviations.
\end{itemize}

\section{Details of Training Setup}
\label{sec:iter-exp}
We provide detailed training configurations used in both the Supervised Fine-Tuning (SFT) and Reinforcement Learning (RL) phases of Unilaw-R1. In the SFT phase, we utilize LoRA to learn the \texttt{<think>...</think>\textbackslash n\textbackslash n<answer>...</answer>} format, with a LoRA rank of 8. In the RL phase, we employ Group Relative Policy Optimization (GRPO) with a group size of 4, which combines a model-based reward signal with policy optimization to ensure legal accuracy and reasoning consistency. The reward signal is generated by a verifier model (Qwen2.5-7B-Instruct) based on legal principles. All our training and test results were performed on machines equipped with 8×96GB NVIDIA H20 GPUs.
Key hyperparameters for both stages are summarized in Table~\ref{tab:training_config_grpo}.

\begin{table}[h]
\small
\centering
\begin{tabular}{lcc}
\toprule
\textbf{Parameter} & \textbf{SFT} & \textbf{RL(GRPO)} \\
\midrule
Batch Size               & 16         & 128 \\
Epochs                   & 5          & 1 \\
Learning Rate            & 1.0e-4       & 1.0e-6 \\
Warmup Ratio             & 0.1       & 0.03 \\
Max Sequence Length      & 4096       & 4096 \\
Gradient Accumulation    & 8          & 4 \\
Optimizer                & AdamW      & AdamW \\
Weight Decay             & 0.01       & 0.01 \\
LR Scheduler             & Cosine     & Cosine \\
Evaluation Interval      & 500 steps  & 10 steps \\
Reward Signal    & –          & Acc \& Fmt \& Legal \\
Reward Granularity       & –          & Step-level \\
Rollout Temperature    & –          & 1.0 \\
Rollout Samples    & –          & 5 \\
KL Coefficient ($\beta$)    & –          & 1.0e-2 \\
Clip Parameter ($\epsilon$)    & –          & 1.0e-6 \\
\bottomrule
\end{tabular}
\caption{Training hyperparameters for SFT and GRPO stages.}
\label{tab:training_config_grpo}
\end{table}

\section{Details of Iterative Inference Setup}
\label{sec:iter-infer}
To enhance the model’s legal reasoning performance through iterative refinement, we adopt a multi-agent setup comprising two collaborative components: an Assessor agent and a Reviser agent. These agents operate in tandem to identify and correct reasoning flaws, enabling a more robust and interpretable inference process.

\subsection{Implementation details}
We employ the Qwen2.5-7B-Instruct model to serve as both the Assessor and Reviser in our iterative inference framework. During the process, we need to evaluate the outcome quality, the InternLM-7B was selected as the outcome reward model (ORM) to computing the chain-level scores. By default, we sample $k=10$ reasoning chains in each iteration, with the decoding temperature parampter fixed at 0.9. The maximum number of iterations is set to 3. We conducted comparative analyses against three distinct methodological approaches:
\begin{itemize}
    \item \textbf{Zero-shot Chain-of-Thought (CoT):} Generates a single reasoning chain per question without subsequent aggregation.
    \item \textbf{Best-of-$k$ Sampling:} Produces multiple candidate chains for each question and selects the optimal output base on maximal ORM score.
    \item \textbf{Majority Vote:} Employs Self-Consistency mechanisms to determine final answers through consensus voting across multiple generated chains.
\end{itemize}

\subsection{The prompt of iterative inference}
\label{sec:iterativeprompt}
\begin{figure*}[h!]
    \centering
    \includegraphics[width=0.98\textwidth]{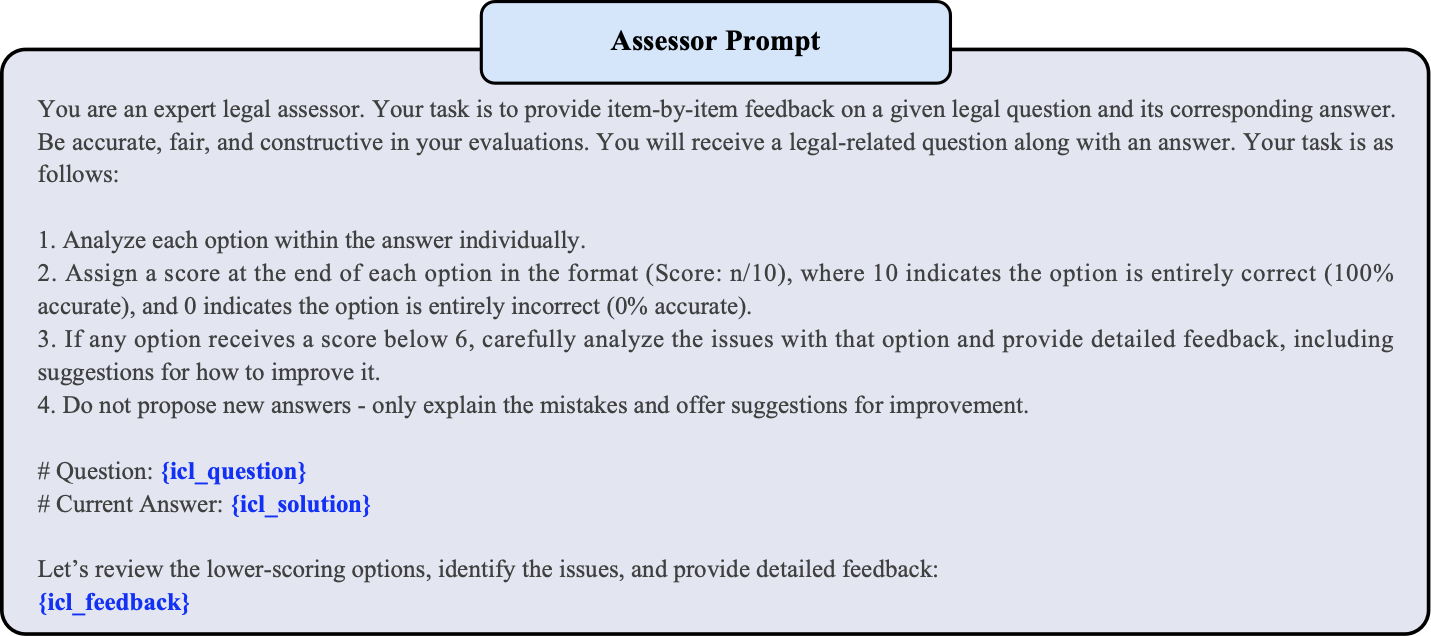} % 将your_image_file.jpg
    \caption{The prompt for assessor the model answer that we used.}
    \label{fig:assessor_prompt}
\end{figure*}

\noindent
\textbf{Assessor Prompt:} The Assessor is tasked with critically evaluating the initial reasoning output from the Unilaw-R1 model. Its prompt is designed to identify potential flaws in logic, incompleteness in option analysis, and inconsistencies with legal principles or cited laws. As illustrated in Figure \ref{fig:assessor_prompt}, the Assessor highlights specific errors or weaknesses and provides structured feedback based on the provided in-context learning question, solution and feedback.

\begin{figure*}[htbp] % figure 
\centering % 
\includegraphics[width=0.98\textwidth]{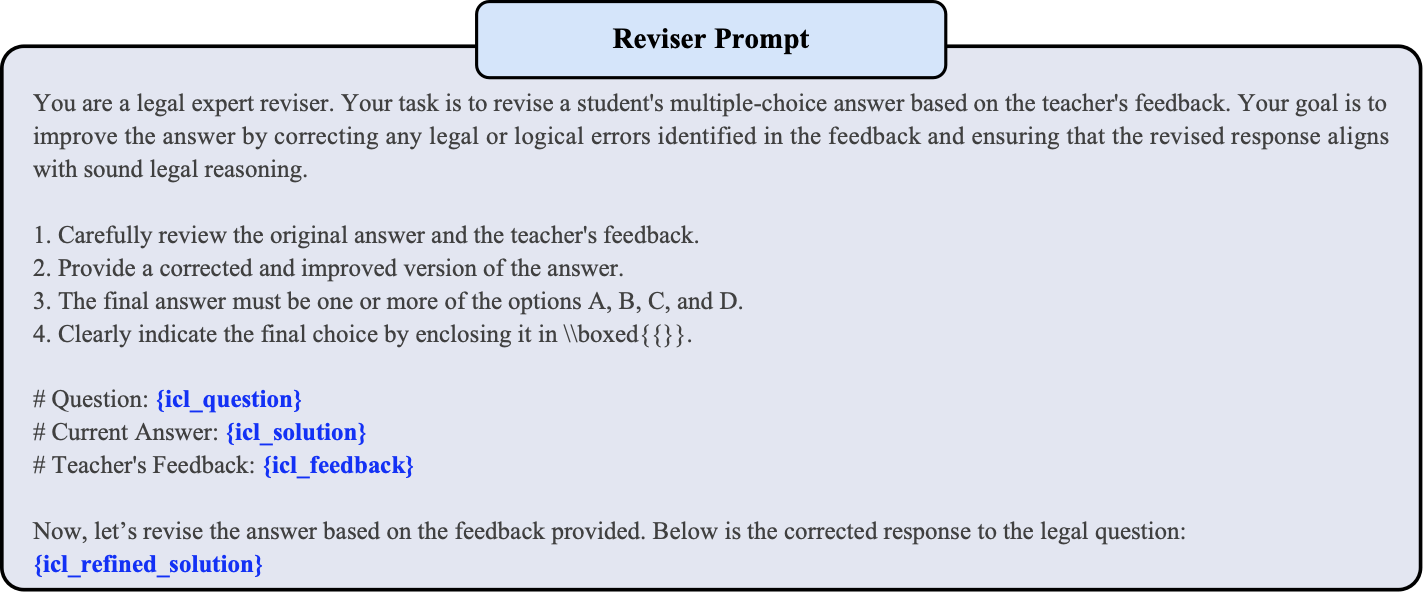}
    \caption{The prompt for revising the model answer that the one-shot in-context learning refined content comes at the end.} % 
    \label{fig:reviser_prompt} % 
\end{figure*}

\noindent
\textbf{Reviser Prompt:} The Reviser then utilizes both the original reasoning and the Assessor's critique to produce an improved version. As shown in Figure \ref{fig:reviser_prompt}, the prompt guides the model to incorporate the Assessor’s feedback while preserving alignment with the legal context and the original question intent. The Reviser ensures that the revised output is not only more accurate but also logically coherent and legally compliant, using the provided one-shot in-context learning example - including question, solution feedback, and the refined solution.

Through multiple rounds of Assessor–Reviser interaction, the system progressively refines its output, achieving higher-quality legal reasoning. This multi-agent collaboration mimics peer-review processes and enhances both the correctness and explainability in legal decision-making.

\subsection{Strategy in instruction model}
We prompt the Qwen2.5-7B-Instruct model to generate explicit reasoning traces enclosed in \texttt{<think>}...\texttt{</think>} tags. Table~\ref{abinstruct} summarizes the model's performance under different inference strategies, including zero-shot Chain-of-Thought (CoT), Best-of-$k$ sampling, Majority Vote, and our proposed Iterative Inference method with varying iteration steps ($Iter=1$ to $Iter=3$). Results on the Unilaw-R1-Eval benchmark demonstrate that Iterative Inference consistently improves performance, achieving the highest overall accuracy (35.9\%) with three iterations.

\begin{table}[htpb]
\vspace{10pt}
\centering
\resizebox{0.48\textwidth}{!}{%
\renewcommand{\arraystretch}{1.1} % 
\begin{tabular}{@{}lcccc@{}}
\toprule[2pt]
\multirow{2}{*}{\textbf{Method}} & \multicolumn{1}{c}{\textbf{SC}} & \multicolumn{2}{c}{\textbf{MC}} & \multicolumn{1}{c}{\textbf{Avg.}} \\
\cmidrule(r){2-2} \cmidrule(lr){3-4} \cmidrule(l){5-5}
    & Acc.(\%)                   & Acc.(\%)         & F1    & Acc.(\%)  \\
\midrule[1pt]
Zero-shot CoT               & 43.2          & 14.2         & 52.4   & 29.9  \\
Best-of-$k$ ($k=10$)            & 52.1          & 10.2         & 56.0    & 32.5 \\
Majority Vote               & 51.9          & 15.0         & 60.2   & 34.6  \\
\midrule[1pt]
Iterative Infer. ($Iter=1$)   & 53.1          & 15.8     & 61.2    & 35.6 \\
Iterative Infer. ($Iter=2$)   & 52.2          & \textbf{17.3}         & \textbf{63.1}   & 35.8  \\
Iterative Infer. ($Iter=3$)   & \textbf{53.3}    & 16.1   & 61.8  & \textbf{35.9}   \\
\bottomrule[2pt]
\end{tabular}
}
\caption{Performance comparison of Qwen2.5-7B-Instruct with different inference methods on the Unilaw-R1-Eval benchmark.}
\label{abinstruct}
\end{table}

\end{CJK}
\end{document}